# Scaling Up Decentralized MDPs Through Heuristic Search

**Jilles S. Dibangoye**
INRIA
Loria, Campus Scientifique - BP 239
54506 Vandœuvre-lès-Nancy, France
*jilles.dibangoye@inria.fr*

**Christopher Amato**
Computer Science and AI Laboratory
Massachusetts Institute of Technology
Cambridge, MA 02139, USA
*camato@csail.mit.edu*

**Arnaud Doniec**
Université Lille Nord de France
Mines Douai, Département IA
F-59500 Douai, France
*arnaud.doniec@mines-douai.fr*

## Abstract

Decentralized partially observable Markov decision processes (Dec-POMDPs) are rich models for cooperative decision-making under uncertainty, but are often intractable to solve optimally (NEXP-complete). The transition and observation independent Dec-MDP is a general subclass that has been shown to have complexity in NP, but optimal algorithms for this subclass are still inefficient in practice. In this paper, we first provide an updated proof that an optimal policy does not depend on the histories of the agents, but only the local observations. We then present a new algorithm based on heuristic search that is able to expand search nodes by using constraint optimization. We show experimental results comparing our approach with the state-of-the-art Dec-MDP and Dec-POMDP solvers. These results show a reduction in computation time and an increase in scalability by multiple orders of magnitude in a number of benchmarks.

## 1 Introduction

There has been substantial progress on algorithms for multiagent sequential decision making represented as decentralized partially observable Markov decision processes (Dec-POMDPs) [18, 20, 7, 2]. Algorithms that are able to exploit domain structure when it is present have been particularly successful [24, 1, 25]. Unfortunately, because the general Dec-POMDP problem is NEXP-complete [8], even these methods cannot solve moderately sized problems optimally.

The decentralized Markov decision process (Dec-MDP) with independent transitions and observations represents a general subclass of Dec-POMDPs that has complexity in NP rather than NEXP [5]. A few algorithms for solving this Dec-MDP subclass have been recently proposed [5, 21, 22]. While these approaches can often solve much larger problems than Dec-POMDP methods, they cannot solve truly large problems or those with more than 2 agents.

In this paper, we present a novel algorithm for optimally solving Dec-MDPs with independent transitions and observations that combines heuristic search and constraint optimization. We show that one can cast any Dec-MDP with independent transitions and observations as a continuous deterministic MDP where states are probability distributions over states in the original Dec-MDP, which we call state occupancy distributions. This allows us to adapt continuous MDP techniques [6, 4] to solve decentralized MDPs. Following this insight, we designed an algorithm where the state occupancy exploration is performed similarly to learning real-time A$^*$ [13] and the policy selection is in accordance with decentralized POMDP techniques [10, 15]. The result is an approach that is able to leverage problem structure through heuristics, limiting the space of policies that are explored by bounding their value and efficiently generating policies with the use of constraint optimization. This algorithm (termed Markov policy search or MPS), is shown to be a much more efficient algorithm than any other approach that can be used in Dec-MDPs with independent transitions and observations.

The remainder of this paper is organized as follows. First, we provide some motivating examples utilizing properties of Dec-MDPs with independent transitions and observations. Next, we describe the Dec-MDP framework and discuss the related work. We then present theoretical results, showing that the optimal policy for Dec-MDPs with independent transitions and observations does not depend on the agent histories. While this has been proven before, we offer a more general proof that permits additional insights. Next, we describe the decentralized Markov policy search algorithm, which combines constraint optimization and heuristic search to more efficiently produce optimal solutions for Dec-MDPs with independent transitions and observations. Finally, we present an empirical evaluation of this algorithm with respect to the state-of-the-art solvers that apply in decentralized MDPs, showing the ability to solve problems that are multiple orders of magnitude larger and those that include up to 10 agents.

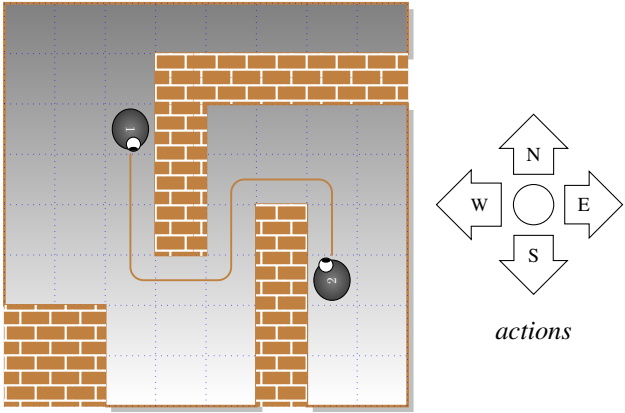

Figure 1: A meeting-grid under uncertainty scenario on a $8 \times 8$ grid, inspired from Seuken and Zilberstein [24].

## 2 Motivating examples

To illustrate the characteristics of decentralized partially observable Markov decision processes (Dec-POMDPs) that we are interested in, consider a simple two-agent "meeting-in-a-grid under uncertainty" domain in Figure 1. In this scenario, two agents want to meet as soon as possible on a two-dimensional grid. In this world, each agent's possible actions include moving north, south, west, east and staying in the same place. The actions of a given agent do not affect the other agents. After taking an action, each agent can sense some information, which in this case corresponds to its own location. Here, each agent's own partial information is insufficient to determine the global state of the world. This is mainly because agents are not permitted to explicitly communicate their local locations with each other. However, if this (instantaneous and noise-free) communication were allowed the agents' partial information together would reveal the true state of the world, (i.e., the agents' joint location). It is the presence of this *joint full observability* property that differentiates Dec-MDPs from Dec-POMDPs.

More generally, in partially observable models including Dec-POMDPs, the agents' partial information together can map to multiple different states of the world. As a consequence, decisions in such models depend on the entire past histories of actions and observations that the agents ever experienced. In the meeting-in-a-grid under uncertainty problem, since both transitions and observations are not affected by the other agents, each agent's decision depends only on its last piece of partial information, (i.e., the agent's own location) [5]. These characteristics appear in many real-world applications including:

**Mars exploration rovers.** The meeting-in-a-grid domain was motivated by a real problem of controlling the operation of multiple space exploration rovers, such as the ones used by NASA to explore the surface of Mars [27].

**Distributed sensor net surveillance.** The sensor net domain [17], where a team of stationary or moveable UAVs, satellites, or other sensors must coordinate to track targets while sensors have independent transitions and observations, is particularly suited to our model.

**Distributed smart-grid domains.** This application aims at finding the optimal schedules and amounts of generated power for a collection of generating units, given demands, and operational constraints over a time horizon.

## 3 Background and Related Work

In this section, we review the decentralized MDP model, the assumptions of transition and observation independence, the associated notation, and related work.

**Definition 1 (The decentralized MDP)** *A $n$-agent decentralized MDP $(S, A, p, r)$ consists of:*

- *A finite set $S = Z^1 \times Z^2 \times \cdots Z^n$ of states $s = \langle z^1, z^2, \cdots, z^n \rangle$, where $Z^i$ denotes the set of local observations $z^i$ of agent $i = 1, 2, \ldots, n$.*

- *A finite set $A = A^1 \times A^2 \times \cdots A^n$ of joint actions $a = \langle a^1, a^2, \cdots, a^n \rangle$, where $A^i$ is the set of local actions $a^i$ of agent $i = 1, 2, \ldots, n$.*

- *A transition function $p(s, a, s')$, which denotes the probability of transiting from state $s = \langle z^1, z^2, \ldots, z^n \rangle$ to state $s' = \langle z'^1, z'^2, \ldots, z'^n \rangle$ when taking joint action $a = \langle a^1, a^2, \ldots, a^n \rangle$*

- *A reward function $r \colon S \times A \mapsto \mathbb{R}$, where $r(s, a)$ denotes the reward received when executing joint action $a$ in state $s$*

As noted above, decentralized MDPs are distinguished by the state being *jointly fully observable*. This property ensures that the global state would be known if all agents shared their observations at a given step (i.e., there is no external uncertainty in the problem) and follows trivially from the definition of states as observations for each agent. The Dec-MDP is parameterized by the initial state distribution $\eta_0$. When the agents operate over a bounded number of steps (typically referred to as the problem horizon) $T$, the model is referred to as a finite-horizon decentralized MDP. Solving a decentralized MDP for a given planning horizon $T$ and start state distribution $\eta_0$ can be seen as finding $n$ individual *policies* that maximize the expected cumulative reward over the steps of the problem.

## 3.1 Additional Assumptions

We are interested in decentralized MDPs that exhibit two properties. The first is the transition independence assumption where the local observation of each agent depends only on its previous local observation and the local action taken by that agent.

**Definition 2 (The transition independent assumption)**
*An $n$-agent decentralized MDP is said to be transition independent if there exists local transition functions $p^1\colon Z^1 \times A^1 \times Z^1 \mapsto [0,1]$, $p^2\colon Z^2 \times A^2 \times Z^2 \mapsto [0,1]$, ..., $p^n\colon Z^n \times A^n \times Z^n \mapsto [0,1]$ such that*

$$p(s, a, s') = \prod_{i=1,\ldots,n} p^i(z^i, a^i, z'^i),$$

*where $s = \langle z^1, z^2, \ldots, z^n \rangle$ and $s' = \langle z'^1, z'^2, \ldots, z'^n \rangle$ and $a = \langle a^1, a^2, \ldots, a^n \rangle$.*

We also implicitly assume observation independence, which states that the observation function of each agent does not depend on the dynamics of the other agents. That is, $P(z'^1, z'^2, \ldots, z'^n | s, a^1, a^2, \ldots, a^n) = \times_i P(z'^i | s, a^i)$. Because we are assuming a Dec-MDP with the state factored into local observations then this becomes the same as transition independence: $\prod_i P(z'^i | z^i, a^i)$.

## 3.2 Preliminary Definitions and Notations

The goal of solving a Dec-MDP is to find a decentralized deterministic joint policy $\pi = \langle \pi^1, \ldots, \pi^n \rangle$. An individual policy $\pi^i$ is a sequence of decision rules $\pi^i = \langle \sigma_0^i, \ldots, \sigma_{T-1}^i \rangle$. In addition, we call decentralized decision rule $\sigma_\tau$ at time $\tau$ an $n$-tuple of decision rules $(\sigma_\tau^1, \ldots, \sigma_\tau^n)$, for $\tau = 0, 1, \ldots, T-1$. In this paper, we distinguish between history-dependent and Markov decision rules.

Each *history-dependent decision rule* $\sigma_\tau^i$ at time $\tau$ maps from $\tau$-step local action-observation histories $h_\tau^i = \langle a_0^i, z_1^i, \ldots, a_{\tau-1}^i, z_\tau^i \rangle$ to local actions: $\sigma_\tau^i(h_\tau^i) = a_\tau^i$, for $\tau = 0, 1, \ldots, T-1$. A sequence of history-dependent decision rules defines a history-dependent policy.

In contrast, each *Markov decision rule* $\sigma_\tau^i$ at time $\tau$ maps from local observations $z_\tau^i$ to local actions: $\sigma_\tau^i(z_\tau^i) = a_\tau^i$, for $\tau = 0, 1, \ldots, T-1$. A sequence of Markov decision rules defines a Markov policy. Moreover, it is worth noticing that decentralized Markov policies are exponentially smaller than decentralized history-dependent ones.

The *state occupancy* is another important notion in this paper. The $\tau$-th state occupancy of a system under the control of a decentralized Markov policy $\langle \sigma_0, \sigma_1, \cdots, \sigma_{\tau-1} \rangle$, denoted $\sigma_{0:\tau-1}$, and starting at $\eta_0$ is given by: $\eta_\tau(s) = P(s | \sigma_{0:\tau-1}, \eta_0)$, for all $\tau \geq 1$. Moreover, the current state occupancy $\eta_\tau$ depends on the past decentralized Markov policy $\sigma_{0:\tau-1}$ only through previous state occupancy $\eta_{\tau-1}$ and decentralized Markov decision rule $\sigma_{\tau-1}$. That is, $\eta_\tau(s') = \sum_s p(s, \sigma_{\tau-1}(s), s') \cdot \eta_{\tau-1}(s)$, for all $\tau \geq 1$. Following [11], this update-rule is denoted $\eta_\tau = \chi_\tau(\eta_{\tau-1}, \sigma_{\tau-1})$ for the sake of simplicity. We also denote $\triangle_\tau$ the state occupancy space at the $\tau$-th horizon, that is the standard $|S|$-dimensional simplex.

**Distinction with belief states.** The state occupancy may be thought of as a belief state, but there are differences. Formally, a belief state $b_\tau$ is given by $b_\tau(s) = P(s | h_\tau, \sigma_{0:\tau-1}, \eta_0)$, for all $\tau \geq 1$ [3]. That is, in belief states, the information agents have about states is typically conditioned on a single joint action-observation history $h_\tau$. From the total probability property, we then have that $\eta_\tau(s) = \sum_{h_\tau} P(s | h_\tau, \sigma_{0:\tau-1}, \eta_0) \cdot P(h_\tau | \sigma_{0:\tau-1}, \eta_0)$. Overall, the $\tau$-th state occupancy summarizes all the information about the world states contained in all belief states at horizon $\tau$. In other words, the doubly exponentially joint action-observation histories are summarized in a single state occupancy that does not make use of local information.

## 3.3 Related Work

In this section, we focus on approaches for solving Dec-MDPs with independent transitions and observations as well as other relevant solution methods. For a thorough introduction to solution methods in Dec-POMDPs, the reader can refer to [24, 20, 7, 2].

Becker et al. [5] were the first to describe the transition and observation independent Dec-MDP subclass and solve it optimally. Their approach, called the coverage set algorithm, consists of three main steps. First, sets of augmented MDPs are created which incorporate the joint reward into local reward functions for each agent. Then, all best responses for any of the other agent policies are found using these augmented MDPs. Finally, the joint policy that has the highest value from all agents' best responses is returned. While this algorithm is optimal, it keeps track of the complete set of policy candidates for each agent, requiring a large amount of time and memory.

Petrik and Zilberstein [22] reformulated the coverage set algorithm as a bilinear program, thereby allowing optimization approaches to be utilized. The bilinear program can be used as an anytime algorithm, providing online bounds on the solution quality at each iteration. The representation is also better able to take advantage of sparse joint reward distributions by representing independent rewards as linear terms and compressing the joint reward matrix. This results in greatly increased efficiency in many cases, but when the agents' rewards often depend on the other agents the bilinear program can still be inefficient due to lack of reward sparsity.

In general Dec-POMDPs, approximate approaches have at-tempted to scale to larger problems and horizons by not generating the full set of policies that may be optimal. These approaches, known as memory-bounded algorithms, were introduced by Seuken and Zilberstein [24] and then successively refined [10, 15]. Memory-bounded algorithms sample forward a bounded number of belief states, and back up (i.e., generate next step policies for) one decentralized history-dependent policy for each belief state. To avoid the explicit enumeration of all possible policies, Kumar and Zilberstein [15] perform the backup by solving a corresponding constraint optimization problem (COP) [9], that represents the decentralized backup. Although, memory-bounded techniques are suboptimal, the decentralized backup can be applied in exact settings as we demonstrate in our algorithm.

More specifically, the decentralized backup can build a horizon-$\tau$ decentralized policy that is maximal with respect to a belief state and horizon-$(\tau + 1)$ policies available for each agent. The associated COP is given by: a set of variables, one for each local observation of each agent; a set of domains, where the domain for the variables corresponding to an agent is the set of horizon-$(\tau + 1)$ policies available for that agent; a set of soft constraints, one for each joint observation. The soft constraint maps assignments to real values. Intuitively, these values represent the expected reward accrued when agents together perceive a given joint observation and follow a given horizon-$(\tau + 1)$ decentralized policy. Since horizon-$\tau$ decentralized policies consist of horizon-$(\tau+1)$ policies, it is easy to see that maximizing the sum of the soft constraints yields a maximal horizon-$\tau$ decentralized policy.

Closer to our model is the ND-POMDP framework [19]. It aims at modeling multiagent teamwork where agents have strong locality of interaction, often through binary interactions. That is, the reward model in such domains is decomposed among sets of agents. There has been a substantial body of work that extend general Dec-POMDP techniques (discussed above) to exploit the locality of interaction [19, 14, 16]. Nair *et* al. [19] introduced the only optimal algorithm for this model, namely the General Optimal Algorithm (GOA). When the domain does not contain binary interactions, there is no reason to expect GOA to outperform general Dec-POMDP algorithms, as all methods use similar strategies in selecting policy candidates. However, when the domain contains primarily binary interactions (or more generally when each agent's rewards are not dependent on many other agents), GOA is likely to outperform general Dec-POMDP algorithms.

It is worth noting that ND-POMDPs and transition and observation independent Dec-MDPs make the same assumptions about transition and observation independence, but make different assumptions about the reward model and partial observability. More specifically, ND-POMDPs assume the reward can be decomposed into the sum of local reward models for sets of agents, while the reward model for transition and observation independent Dec-MDPs is more general, allowing global rewards for all agents (i.e., considering all agents to be in one set). Dec-MDPs assume that the state is jointly fully observable (i.e., that the state is fully determined by the combination of local observations of all agents), while ND-POMDPs do not make this limiting assumption. Both models therefore make different assumptions to address complexity and the choice of model depends on which assumptions best match the domain being solved.

## 4 Theoretical Properties

In this section, we demonstrate the main theoretical results of this paper.

### 4.1 Optimal Policies

A decentralized MDP solver aims to calculate an optimal decentralized policy $\pi^*$ that maximizes the expected cumulative reward:

$$\pi^* = \arg\max_\pi E[\sum_{\tau=0}^{T-1} r(s_\tau, a_\tau)|\pi, \eta_0]. \quad (1)$$

The following theorem proves that decentralized Markov policies yield the optimal performance in decentralized MDPs with independent transitions and observations. Goldman *et* al. [12] established the optimality of Markov policy for an agent under the assumption that the other agents choose Markov policies. Here, we state the optimality of Markov policies for an agent no matter what its teammates' policies are. We also construct the proof in a manner that more directly relates policies to values (rather than information sets). This may be more clear to some readers.

**Theorem 1 (Optimality of decentralized Markov policies)** *In Dec-MDPs with independent transitions and observations, optimal policies for each agent depend only on the local state and not on agent histories.*

**Proof** Without loss of generality, we construct a proof by induction for two agents, 1 and 2, from agent 1's perspective. We first show that in the last step of the problem, agent 1's policy does not depend on its local history.

Agent 1's local policy on the last step is: $\sigma_{T-1}^{1*}(h_{T-1}^1) = \arg\max_{a^1} \sum_{h_{T-1}^2} P(h_{T-1}^2|h_{T-1}^1) \cdot R(s, a^1, \sigma_{T-1}^2(h_{T-1}^2))$, which chooses a local action to maximize value based on the possible local histories of agent 2 and resulting states of the system $s = \langle z_{T-1}^1, z_{T-1}^2 \rangle$.

Based on transition and observation independence and the use of decentralized policies, it can be shown that $P(h_{T-1}^2|h_{T-1}^1) = P(h_{T-1}^2)$. Due to space limitations, we do not include full proof of this claim. Intuitively it holds because each agent does not receive any information about the other agents' local histories due to transition independence. Therefore, we can represent agent 1's policy on the last step as $\sigma^{1^*}_{T-1}(h^1_{T-1}) = \arg\max_{a^1} \sum_{h^2_{T-1}} P(h^2_{T-1}) \cdot R(s, a^1, \sigma^2_{T-1}(h^2_{T-1}))$ which no longer depends on the history $h^1_{T-1}$. Therefore, the policy on the last step for either agent does not depend on history.

This allows us to define the value function on the last step as $v_{T-1}(s, \sigma^1_{T-1}(z^1_{T-1}), \sigma^2_{T-1}(z^2_{T-1}))$.

Then for the induction step, we can show that if the policy at step $\tau + 1$ does not depend on history, then the policy at step $\tau$ also does not depend on its local history. Again, we show this from agent 1's perspective.

Agent 1's policy on step $\tau$ can be represented by: $\sigma^{1^*}_\tau(h^1_\tau) = \arg\max_{a^1} \sum_{h^2_\tau} P(h^2_\tau | h^1_\tau) \cdot v_{\tau+1}(s, a^1, \sigma^2_\tau(h^2_\tau))$, where the value function $v_{\tau+1}$ is assumed to not depend on history. We can again show that $P(h^2_\tau | h^1_\tau) = P(h^2_\tau)$ because of transition independence and represent agent 1's policy on step $\tau$ as:

$$\sigma^{1^*}_\tau(h^1_\tau) = \arg\max_{a^1} \sum_{h^2_\tau} P(h^2_\tau) \cdot v_{\tau+1}(s, a^1, \sigma^2_\tau(h^2_\tau))$$

which no longer depends on the local history $h^1_\tau$.

Therefore, the policy of either agent does not depend on local history for any step of the problem. ∎

We now establish the sufficient statistic for the selection of decentralized Markov decision rules.

**Theorem 2 (Sufficient Statistic)** *The state occupancy is a sufficient statistic for decentralized Markov decision rules.*

**Proof** We build upon the proof of the optimality of decentralized Markov policies in Theorem 1. We note that an optimal decentralized Markov policy starting in $\eta_0$ is given by:

$$\pi^* = \arg\max_\pi \sum_\tau \sum_{h_\tau} P(h_\tau | \sigma_{0:\tau-1}, \eta_0) \cdot r(s_\tau, \sigma_\tau[s_\tau])$$

The substitution of $h_\tau$ by $(h_{\tau-1}, a_{\tau-1}, s_\tau)$ plus the sum over all pairs $(h_{\tau-1}, a_{\tau-1})$ yields

$$\pi^* = \arg\max_\pi \sum_\tau \sum_{s_\tau} P(s_\tau | \sigma_{0:\tau-1}, \eta_0) \cdot r(s_\tau, \sigma_\tau[s_\tau]),$$

We denote $\eta^\pi_\tau = P(s_\tau | \sigma_{0:\tau-1}, \eta_0)$ the state occupancy distribution that decentralized Markov policy $\pi$ produced at horizon $\tau$. And hence,

$$\pi^* = \arg\max_\pi \sum_\tau \sum_{s_\tau \in S} \eta^\pi_\tau(s_\tau) \cdot r(s_\tau, \sigma_\tau[s_\tau])$$

So, state occupancy $\eta^\pi_\tau$ summarizes all possible joint action-observation histories $h_\tau$ decentralized Markov policy $\pi$ produced at horizon $\tau$ for the estimate of joint decision rule $\sigma_\tau$. Thus, the state occupancy is a sufficient statistic for decentralized Markov decision rules since their estimates depend only upon a state occupancy, and no longer on all possible joint observation-histories. ∎

States, belief states, and multi-agent belief states are all sufficient to select directly actions for MDPs, POMDPs, and decentralized POMDPs, respectively. This is mainly because all these statistics summarize the information about the world states from a single agent perspective. The state occupancy, instead, summarizes the information about the world states from the perspective of a team of agents that are constrained to execute their policies independently from each other. In such a setting, joint actions cannot be selected independently, instead, they are selected jointly through decentralized Markov decision rules.

### 4.2 Optimality Criterion

This section presents the optimality criterion based on the policy value functions.

We first define *the $\tau$-th expected immediate reward function* $r_\tau(\cdot, \sigma_\tau) \colon \triangle_\tau \mapsto \mathbb{R}$ that is given by $r_\tau(\eta_\tau, \sigma_\tau) = E_{s \sim \eta_\tau}[r(s, \sigma_\tau[s])]$. This quantity denotes the immediate reward of taking decision rule $\sigma_\tau$ when the system is in state occupancy $\eta_\tau$ at the $\tau$-th time step.

Let $v_\pi(\eta_0)$ represent *the expected total reward* over the decision making horizon if policy $\pi$ is used and the system is in state occupancy $\eta_0$ at the first time step $\tau = 0$. For $\pi$ in the space of decentralized Markov policies, the expected total reward is given by:

$$v_\pi(\eta_0) \equiv E_{(\eta_1, \cdots, \eta_{T-1})}\left[\sum_{\tau=0}^{T-1} r_\tau(\eta_\tau, \sigma_\tau) \mid \eta_0, \pi\right]$$

We say that a decentralized Markov policy $\pi^*$ is *optimal* under the total reward criterion whenever $v_{\pi^*}(\eta_0) \geq v_\pi(\eta_0)$ for all decentralized Markov policies $\pi$.

Following the Bellman principle of optimality [23], one can separate the problem of finding the optimal policy $\pi^*$ into simpler subproblems. Each of these subproblems consists of finding policies $\sigma_{\tau:T-1}$ that are optimal for all $\tau = 0, \cdots, T-1$. To do so, we then define the *$\tau$-th value function* $v_{\sigma_{\tau:T-1}} \colon \triangle_\tau \mapsto \mathbb{R}$ under the control of decentralized Markov policy $\sigma_{\tau:T-1}$ as follows:

$$v_{\sigma_{\tau:T-1}}(\eta_\tau) = r_\tau(\eta_\tau, \sigma_\tau) + v_{\sigma_{\tau+1:T-1}}(\chi_{\tau+1}(\eta_\tau, \sigma_\tau))$$

where quantity $v_{\sigma_{\tau:T-1}}(\eta_\tau)$ denotes the expected sum of rewards attained by starting in state occupancy $\eta_\tau$, taking one joint action according to $\sigma_\tau$, taking the next joint action according to $\sigma_{\tau+1}$, and so on. We slightly abuse notation and write the $\tau$-th value function under the control of an "unknown" decentralized Markov policy $\sigma_{\tau:T-1}$ using $v_\tau \colon \triangle_\tau \mapsto \mathbb{R}$.

We further denote $\mathscr{V}_\tau$ to be the space of bounded value functions at the $\tau$-th horizon. For each $v_{\tau+1} \in \mathscr{V}_{\tau+1}$, and decentralized Markov decision rule $\sigma_\tau$, we define the *linear transformation* $\mathscr{L}_{\sigma_\tau} \colon \mathscr{V}_{\tau+1} \mapsto \mathscr{V}_\tau$ by

$$[\mathscr{L}_{\sigma_\tau} v_{\tau+1}](\eta_\tau) = r_\tau(\eta_\tau, \sigma_\tau) + v_{\tau+1}(\chi_{\tau+1}(\eta_\tau, \sigma_\tau)).$$

As such, the $\tau$-th value function $v_\tau$ can be built from a $(\tau + 1)$-th value function $v_{\tau+1}$ as follows:

$$\begin{aligned} v_\tau(\eta_\tau) &= \max_{\sigma_\tau} \ [\mathscr{L}_{\sigma_\tau} v_{\tau+1}](\eta_\tau), \\ v_T(\eta_T) &= 0. \end{aligned} \quad (2)$$

In our setting, Equations (2) denote the optimality equations. It is worth noting that the decentralized Markov policy solution $\pi = \langle \sigma_0, \cdots, \sigma_{T-1} \rangle$ of the optimality equations is greedy with respect to value functions $v_0, \ldots, v_{T-1}$.

## 5 Markov Policy Search

In this section, we compute optimal decentralized Markov policy $\langle \sigma_0^*, \cdots, \sigma_{T-1}^* \rangle$ given initial state occupancy $\eta_0$ and planning horizon $T$. Note that while state occupancies are used to calculate heuristics in this algorithm, the final choices at each step do not depend on the state occupancies. That is, the result is a nonstationary policy for each agent mapping local observations to actions at each step.

We cast decentralized MDPs $(S, A, p, r)$ as continuous and deterministic MDPs where: states are state occupancy distributions $\eta_\tau$; actions are decentralized Markov policies $\sigma_\tau$; the update-rules $\chi_\tau(\cdot, \sigma_{\tau-1})$ define transitions; and mappings $r_\tau(\cdot, \sigma_\tau)$ denote the reward function. So, techniques that apply in continuous and deterministic MDPs also apply in decentralized MDPs with independent transitions and observations. For the sake of efficiency, we focus only on optimal techniques that exploit the initial information $\eta_0$.

The *learning real-time A$^*$* (LRTA$^*$) algorithm can be used to solve deterministic MDPs [13]. This approach updates only states that agents actually visit during the planning stage. Therefore, it is suitable for continuous state spaces. Algorithm 1, namely *Markov Policy Search* (MPS), illustrates an adaptation of the LRTA$^*$ algorithm for solving decentralized MDPs with independent transitions and observations. The MPS algorithm relies on lower and upper bounds $\underline{v}_\tau$ and $\bar{v}_\tau$ on the exact value functions for all planning horizons $\tau = 0, \ldots, T - 1$.

We use the following definitions. Q-value functions $\bar{q}_\tau(\eta_\tau, \sigma_\tau)$ denote rewards accrued after taking decision rule $\sigma_\tau$ at state occupancy $\eta_\tau$ and then following the policy defined by upper-bound value functions for the remaining planning horizons. We denote $\Psi_\tau(\eta_\tau) = \{\sigma_\tau\}$ to be the set of all stored decentralized Markov decision rules for state occupancy $\eta_\tau$. Thus, $\bar{v}_\tau(\eta_\tau) = \max_{\sigma_\tau \in \Psi_\tau(\eta_\tau)} \bar{q}_\tau(\eta_\tau, \sigma_\tau)$ represents the upper-bound value at state occupancy $\eta_\tau$. Formally, we have that $\bar{q}_\tau(\eta_\tau, \sigma_\tau) = [\mathscr{L}_{\sigma_\tau} \bar{v}_{\tau+1}](\eta_\tau)$.

Next, we describe two variants of the MPS algorithm. The exhaustive variant replaces states by state occupancy distributions, and actions by decentralized Markov decision rules in the LRTA$^*$ algorithm. The second variant uses a constraint optimization program instead of the memory demanding exhaustive backup operation that both the LRTA$^*$ algorithm and the exhaustive variant use.

### 5.1 The exhaustive variant

The exhaustive variant consists of three major steps: the initialization step (line 1); the backup operation step (line 5); and the update step (lines 6 and 8). It repeats the execution of these steps until convergence $(\bar{v}_0(\eta_0) - \underline{v}_0(\eta_0) \leq \varepsilon)$. At this point, an $\epsilon$-optimal decentralized Markov policy has been found.

---

**Algorithm 1:** The MPS algorithm.

**begin**
1    Initialize bounds $\underline{v}$ and $\bar{v}$.
2    **while** $\bar{v}_0(\eta_0) - \underline{v}_0(\eta_0) > \varepsilon$ **do**
3      MPS-TRIAL$(\eta_0)$

MPS-TRIAL$(\eta_\tau)$ **begin**
4    **while** $\bar{v}_\tau(\eta_\tau) - \underline{v}_\tau(\eta_\tau) > \varepsilon$ **do**
5      $\sigma_{\text{greedy},\tau} \leftarrow \arg\max_{\sigma_\tau} \bar{q}_\tau(\eta_\tau, \sigma_\tau)$
6      Update the upper bound value function.
7      MPS-TRIAL$(\chi_{\tau+1}[\eta_\tau, \sigma_{\text{greedy},\tau}])$
8      Update the lower bound value function.

---

**Initialization.** We initialize lower bound $\underline{v}_\tau$ with the $\tau$-th value function of any decentralized Markov policy, such as a randomly generated policy $\pi_{\text{rand}} = \langle \sigma_{\text{rand},0}, \ldots, \sigma_{\text{rand},T-1} \rangle$, where $\underline{v}_\tau = v_{\sigma_{\text{rand},\tau}, \ldots, \sigma_{\text{rand},T-1}}$. We initialize the upper bound $\bar{v}_\tau$ with the $\tau$-th value function of the underlying MDP. That is, $\pi_{\text{mdp}} = \langle \sigma_{\text{mdp},0}, \ldots, \sigma_{\text{mdp},T-1} \rangle$, where $\bar{v}_\tau = v_{\sigma_{\text{mdp},\tau}, \ldots, \sigma_{\text{mdp},T-1}}$.

**The exhaustive backup operation.** We choose decentralized Markov decision rule $\sigma_{\text{greedy},\tau}$, which yields the highest value $\bar{v}_\tau(\eta_\tau)$ through the explicit enumeration of all possible decentralized Markov decision rules $\sigma_\tau$. We first store all decentralized Markov decision rules $\sigma_\tau$ for each visited state occupancy $\eta_\tau$ together with corresponding values $\bar{q}_\tau(\eta_\tau, \sigma_\tau)$. Hence, the greedy decentralized Markov decision rule $\sigma_{\text{greedy},\tau}$ is $\arg\max_{\sigma_\tau} \bar{q}_\tau(\eta_\tau, \sigma_\tau)$ at state occupancy $\eta_\tau$.

**Update of lower and upper bounds.** We update the lower bound value function based on decentralized Markov policies $\pi_{\text{greedy}} = \langle \sigma_{\text{greedy},0}, \ldots, \sigma_{\text{greedy},T-1} \rangle$ selected at each trial. If $\pi_{\text{greedy}}$ yields a value higher than that of the current lower bound, $\underline{v}_0(\eta_0) < v_{\pi_{\text{greedy}}}(\eta_0)$, we set $\underline{v}_\tau = v_{\sigma_{\text{greedy},\tau}, \ldots, \sigma_{\text{greedy},T-1}}$ for $\tau = 0, \ldots, T - 1$, otherwise we leave the lower bound unchanged. We update the upper bound value function based on decentralized Markov decision rules $\sigma_{\text{greedy},\tau}$ and the $(\tau + 1)$-th upper-bound value function $\bar{v}_{\tau+1}$, as follows $\bar{v}_\tau(\eta_\tau) = [\mathscr{L}_{\sigma_{\text{greedy},\tau}} \bar{v}_{\tau+1}](\eta_\tau)$.

**Theoretical guarantees.** The exhaustive variant of MPS yields both advantages and drawbacks. On the one hand, it inherits the theoretical guarantees from the LRTA$^*$ al-gorithm. In particular, it terminates with a decentralized Markov policy within $\varepsilon = \bar{v}_0(\eta_0) - \underline{v}_0(\eta_0)$ of the optimal decentralized Markov policy. Indeed, the upper bound value functions $\bar{v}_\tau$ never underestimate the exact value at any state occupancy $\eta_\tau$. This is because we update the upper bound value at each state occupancy based upon a greedy decision rule for this state occupancy. On the other hand, the exhaustive variant algorithm requires the exhaustive enumeration of all possible decentralized Markov decision rules at each backup step (Algorithm 1, line 5). In MDP techniques, the exhaustive enumeration is not prohibitive since the action space is often manageable. In decentralized MDP planning, however, the space of all decentralized Markov decision rules increases exponentially with increasing observations and agents. As such, the exhaustive variant can scale only to problems with a moderate number of observations (local states) and two agents.

### 5.2 The constraint optimization formulation

To overcome the memory limitation of the exhaustive variant, we use constraint optimization instead of the exhaustive backup operation. More precisely, our constraint optimization program returns a greedy decentralized Markov decision rule $\sigma_{\text{greedy},\tau}$ for each state occupancy $\eta_\tau$ visited, but without performing the exhaustive enumeration.

In our constraint optimization formulation, variables are associated with decision rules $\sigma_\tau^i(z^i)$ for all agents $i = 1,\ldots,n$ and all local observations $z^i \in Z^i$. The domain for each variable $\sigma_\tau^i(z^i)$ is action space $A^i$. For each state $s \in S$, we associated a single soft constraint $c_\tau(s,\cdot) \colon A \mapsto \mathbb{R}$. Each of these assigns a value $c_\tau(s,a) = r(s,a) + \sum_{s'} p(s,a,s') \cdot v_{\sigma_{\text{mdp},\tau+1},\ldots,\sigma_{\text{mdp},T-1}}(s')$ to each joint action $a \in A$. Value $c_\tau(s,a)$ denotes the reward accrued at horizon $\tau$ when taking joint action $a$ in state $s$ and then following the underlying MDP joint policy for the remaining planning horizons. For each decentralized Markov decision rule $\sigma_\tau \in \Psi_\tau(\eta_\tau)$, we also associate a single soft constraint $g_\tau(\cdot)$. Each of these assigns value $g_\tau(\sigma_\tau) = \bar{q}_\tau(\eta_\tau,\sigma_\tau) - \bar{q}_{\text{mdp}}(\eta_\tau,\sigma_\tau)$, where $\bar{q}_{\text{mdp}}(\eta_\tau,\sigma_\tau) = [\mathscr{L}_{\sigma_\tau} v_{\sigma_{\text{mdp},\tau+1},\ldots,\sigma_{\text{mdp},T-1}}](\eta_\tau)$. The objective of our constraint optimization model is to find an assignment $\sigma_{\text{greedy},\tau}$ of actions $a^i$ to variables $\sigma_\tau^i(z^i)$ such that the aggregate value is maximized. Stated formally, we wish to find $\sigma_{\text{greedy},\tau} = \arg\max_{\sigma_\tau} g_\tau(\sigma_\tau) + \sum_s \eta_\tau(s) c_\tau(s,\sigma_\tau(s))$. To better understand our constraint optimization program, note that by the definition of mapping $\bar{q}_{\text{mdp}}$ we have that $\sum_s \eta_\tau(s) \cdot c_\tau(s,\sigma_\tau(s)) = \bar{q}_{\text{mdp}}(\eta_\tau,\sigma_\tau)$. Hence, if we use $\bar{q}_{\text{mdp}}(\eta_\tau,\sigma_\tau)$ instead of $\sum_s \eta_\tau(s) \cdot c_\tau(s,\sigma_\tau(s))$, we get $\sigma_{\text{greedy},\tau} = \arg\max_{\sigma_\tau} \bar{q}_\tau(\eta_\tau,\sigma_\tau)$. Thus, our constraint optimization program returns a decentralized Markov decision rule with the highest upper-bound value. Techniques that solve our constraint optimization formulation abound in the literature of constraint programming [9], allowing many different approaches to be utilized.

**Theoretical guarantees.** The constraint optimization variant yields the same guarantees as the exhaustive variant without the major drawback of exhaustive enumeration of all decentralized Markov decision rules. Instead, it uses a constraint optimization formulation that returns a greedy decentralized Markov decision rule, which will often be much more efficient than exhaustive enumeration. And hence, we retain the property that stopping the algorithm at any time, the solution is within $\varepsilon = \bar{v}_0(\eta_0) - \underline{v}_0(\eta_0)$ of an optimal decentralized Markov policy.

**Comparison to COP based algorithms.** There is a rich body of work that replaces the exhaustive backup operation by a constraint optimization formulation in decentralized control settings [15, 14, 19]. These constraint optimization programs compute a decentralized history-dependent policy for a given belief state. While MPS also takes advantage of a constraint optimization formulation, it remains fundamentally different. The difference lies in both the COP formulation and the heuristic search. In existing COP based algorithms for decentralized control, authors try to find the best assignment of sub-policies to histories. Instead, in our case the COP formulation aims at mapping local observations to local actions. This provides considerable memory and time savings. Moreover, existing algorithms proceed by backing up policies in a backward direction (i.e., from last step to first) using a set pre-selected belief states. In contrast, the MPS algorithm proceeds forward, expanding the state occupancy distributions and selecting greedily decision rules. Finally, the MPS algorithm returns an optimal solution, whereas other COP based algorithms for Dec-POMDPs return only locally optimal solutions [15, 14]. Approximate solutions are returned by the other algorithms because they plan over (centralized) belief states, which do not constitute a sufficient statistic for Dec-POMDPs (or Dec-MDPs).

## 6 Empirical Evaluations

We evaluated our algorithm using several benchmarks from the decentralized MDP literature. For each benchmark, we compared our algorithms with state-of-the-art algorithms for solving Dec-MDPs and Dec-POMDPs. Note that we do not compare with ND-POMDP methods. Since our benchmarks allow all agents to interact with all teammates at all times, there is no reason to expect the optimal ND-POMDP method (GOA [19]) to outperform the algorithms presented here. We report on each benchmark the optimal value $v_0(\eta_0)$ together with the running time in seconds for different planning horizons.

The MPS variants were run on a Mac OSX machine with 2.4GHz Dual-Core Intel and 2GB of RAM available. We solved the constraint optimization problems using the aolib library[1]. The bilinear programming approach (listed as BLP) was run on a 2.8GHz Quad-Core Intel Mac with 2GB of RAM with a time limit of 3 hours. We used the best available version of the bilinear program approach which was the iterative best response version with standard parameters. This is a generic solution method which does not perform as well as the more specialized approaches in [22], but we do not expect results to differ by more than a single order of magnitude. We do not compare to the coverage set algorithm because the bilinear programming methods have been shown to be more efficient for all available test problems.

---
[1] The aolib library is available at the following website:

| $T$ | $v_0(\eta_0)$ | ICE | IPG | BLP | MPS | |
|---|---|---|---|---|---|---|
| | | | | | exh | COP |
| Recycling robot ($|Z|=4, |A|=9$) | | | | | | |
| 50 | 154.94 | 1.27 | - | 8848.7 | 0.016 | 0.5 |
| 60 | 185.71 | 6.00 | - | - | 0.090 | 0.555 |
| 70 | 216.47 | 28.6 | - | - | 0.111 | 0.395 |
| 80 | 247.24 | - | - | - | 0.124 | 0.545 |
| 90 | 278.01 | - | - | - | 0.151 | 0.373 |
| 100 | 308.78 | - | - | - | 0.156 | 0.438 |
| **1000** | **3078.0** | - | - | - | **1.440** | **5.374** |
| Meeting Grid ($|Z|=81, |A|=25$) | | | | | | |
| 2 | 0.0 | 0.00 | 5 | 10.0 | - | 0.030 |
| 3 | 0.13 | 0.02 | 17 | 34.7 | - | 0.110 |
| 4 | 0.43 | 0.37 | 54 | 192.8 | - | 0.114 |
| 5 | 0.89 | 4.38 | 600 | 571.2 | - | 0.131 |
| 6 | 1.49 | - | - | 1160.6 | - | 0.159 |
| 10 | 4.68 | - | - | 3938.5 | - | 0.309 |
| **100** | **94.26** | - | - | - | - | **13.12** |
| **1000** | **994.2** | - | - | - | - | **33.59** |

| $T$ | $v_0(\eta_0)$ | ICE | IPG | BLP | COP |
|---|---|---|---|---|---|
| Meeting on a 8x8 Grid ($|Z|=4096, |A|=25$) | | | | | |
| 5 | 0.0 | 12.55 | - | - | 5.05 |
| **6** | **0.0** | - | - | - | **6.20** |
| 7 | 0.71 | - | - | - | 13.16 |
| 8 | 1.67 | - | - | - | 13.76 |
| 9 | 2.68 | - | - | - | 16.94 |
| 10 | 3.68 | - | - | - | 18.66 |
| 20 | 13.68 | - | - | - | 38.37 |
| 30 | 23.68 | - | - | - | 52.39 |
| 40 | 33.68 | - | - | - | 59.70 |
| 50 | 43.68 | - | - | - | 74.28 |
| 100 | 93.68 | - | - | - | 214.73 |
| Navigation (MIT) ($|Z|=7225, |A|=16$) | | | | | |
| 10 | 0.0 | 85.85 | - | - | 47.322 |
| **20** | **0.0** | - | - | - | **321.26** |
| 30 | 14.28 | - | - | - | 180.70 |
| 40 | 34.97 | - | - | - | 400.48 |
| 50 | 54.92 | - | - | - | 1061.94 |
| 100 | 154.93 | - | - | - | 1236.11 |

| $T$ | $v_0(\eta_0)$ | ICE | IPG | BLP | COP |
|---|---|---|---|---|---|
| Navigation (ISR) ($|Z|=8100, |A|=16$) | | | | | |
| 2 | 0.0 | 0.71 | - | 3225.8 | 2.39 |
| 3 | 0.0 | 6.50 | - | - | 3.19 |
| **4** | **0.0** | - | - | - | **4.31** |
| 5 | 0.38 | - | - | - | 13.43 |
| 10 | 6.47 | - | - | - | 54.16 |
| 50 | 83.02 | - | - | - | 194.66 |
| 100 | 182.54 | - | - | - | 1294.28 |
| Navigation (PENTAGON) ($|Z|=9801, |A|=16$) | | | | | |
| 2 | 0.0 | 0.99 | - | 4915.1 | 1.31 |
| 3 | 0.0 | 6.01 | - | - | 5.11 |
| **4** | **0.0** | - | - | - | **8.75** |
| 5 | 0.38 | - | - | - | 13.81 |
| 10 | 4.82 | - | - | - | 62.89 |
| 20 | 19.73 | - | - | - | 129.48 |
| 30 | 39.18 | - | - | - | 209.11 |
| 40 | 57.75 | - | - | - | 276.20 |
| 50 | 76.39 | - | - | - | 1033.70 |

Table 1: Experimental results for the *COP* and *exh.* variants of MPS as well as GMAA$^*$-ICE (labeled ICE), IPG, and BLP.

We provide values for the exhaustive variant, *exh*, on small problems and constraint optimization formulation, *COP*, for all problems. We tested our algorithms on six benchmarks: recycling robot, meeting-in-a-grid 3x3 and 8x8; and navigation problems[2]. These are the largest and hardest benchmarks we could find in the literature. We compare our algorithms with: GMAA$^*$-ICE [25], IPG [1], and BLP. The GMAA$^*$-ICE heuristic search consistently outperforms other generic exact solvers such as (G)MAA$^*$ [25]. The IPG algorithm is a competitive alternative to the GMAA$^*$ approach and performs well on problems with reduced reachability [1]. Results for GMAA$^*$-ICE were provided by Matthijs Spaan and as such were conducted on a different machine. Similarly, results for IPG were collected on different machine. As a result, the timing results for GMAA$^*$-ICE and IPG are not directly comparable to the other methods, but are likely to only differ by a small constant factor from those that would be obtained on our test machine.

The results can be seen in Table 1. In all benchmarks, the *COP* variant of MPS outperforms the other algorithms. The results show that the *COP* variant produces the optimal policies in much less time for all tested benchmarks. For example, in the meeting in a 3x3 grid problem for $T=5$: the *COP* variant computed the optimal policies approximately 33, 4358 and 4580 times faster than the GMAA$^*$-ICE, BLP and IPG algorithms, respectively. We also note that the *COP* variant is very useful for the medium and large domains. For example, in all large domains, the *exh.* variant ran out of memory while the *COP* variant computed the optimal solutions for horizons up to 100. Yet, the *exh.* variant can compute the optimal solution of small problems faster than the *COP* variant. For instance, in the recycling robot for horizon $T=1000$, the *exh.* variant computed the optimal solution in about 5 times faster than the *COP* variant of the MPS algorithm due to overhead in the constraint optimization formulation and a lack of structure that can be utilized.

There are many different reasons for these results. The MPS algorithm outperforms GMAA$^*$-ICE and IPG mainly because they perform a policy search in the space of decentralized history-dependent policies. Instead, the MPS algorithm performs its policy search in the space of decentralized Markov policies, which is exponentially smaller than that of the decentralized history-dependent policies. The MPS outperforms the BLP algorithm mainly because of the dimension of its solution representation. More specifically, the number of bilinear terms in the BLP approach grows polynomially in the horizon of the problem, causing it to not perform well for large problems and large horizons with tightly coupled reward values.

We continue the evaluation of the MPS algorithm on randomly generated instances with multiple agents. The random instances were built upon the recycling robot problem described in Sutton and Barto [26]. Given $n$ such models, each of which is associated with a single agent, we choose a number of interaction events. An interaction event is a pair of joint states and actions $(s,a)$ where the reward $r(s,a)$

---

http://graphmod.ics.uci.edu/group/aolibWCSP/

[2] All problem definitions are available at the following website: http://users.isr.ist.utl.pt/$\sim$ mtjspaan/decpomdp/

is randomly chosen. This structure ties all agents together since the reward model cannot be decomposed among subgroups of agents. In an effort to provide insight on the degree of interaction among all agents, we distinguish between four classes $\{c_0, c_1, c_2, c_3\}$, each of which depends on the number of interaction events $e$. For each class $c_k$, we randomly choose $e$ such that $e \in [\frac{k}{4}e_{\max}; \frac{k}{4}(1+e_{\max})]$, where $e_{\max}$ denotes the number of joint state and action pairs.

As depicted in Figure 2, the constraint formulation allows us to deal with larger numbers of agents. We calculated optimal value functions for 100 instances of each class, and reported the average computational time. The *COP* variant was able to scale up to 6 agents at horizon 10 in about $5,000$ seconds. We could also produce results for up to 10 agents in about $40,000$ seconds using a more powerful machine. It can also be seen that increasing the number of interaction events on each problem does not substantially increase the amount of time required to solve these problems. This shows that even for dense reward matrices, our approach will continue to perform well. Despite this high running time, MPS is the first generic algorithm that scales to teams of more than two agents without taking advantage of the locality of interaction. For example the BLP algorithm as it currently stands can only solve two-agent problems. Moreover, ND-POMDP techniques exploit the small number of local interactions among agents to scale to multiple agents, but in this problem all agents interact.

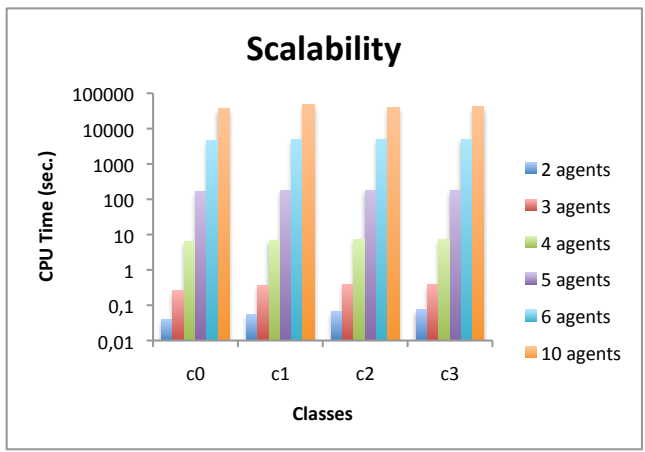

Figure 2: The MPS performance for increasing number of agents at planning horizon $N = 10$ for randomized instances of the recycling robot scenario.

## 7 Conclusion and Future Work

This paper explores new theory and algorithms for solving independent transition and observation Dec-MDPs. We provide a new proof that optimal policies do not depend on agent histories in this subclass, generalizing previous theoretical results. We also describe a novel algorithm that combines heuristic search and constraint optimization to more efficiently produce optimal solutions for this class of problems. This new algorithm, termed learning Markov policy or MPS, was shown to scale up to large problems and planning horizons, reducing computation time by multiple orders of magnitude over previous approaches. We were also able to demonstrate scalability with respect to the number of agents in domains with up to 10 agents. These results show that our approach could be applied to many large and realistic domains.

In the future, we plan to explore extending the MPS algorithm to other classes of problems and larger teams of agents. For instance, we may be able to produce an optimal solution to more general classes of Dec-MDPs or provide approximate results for Dec-POMDPs by extending the idea of an occupancy distribution to those problems. Furthermore, the scalability of our approach to larger numbers of agents is encouraging and we will pursue methods to increase this even further. In particular, we think our approach could help increase the number of agents that interact in conjunction with other structure in the model such as locality of interaction (as in ND-POMDPs) or sparse joint reward matrices (as in bilinear programming approaches).

## 8 Acknowledgements

We would like to thank Frans Oliehoek and the anonymous reviewers for their helpful comments about initial versions of the paper as well as Matthijs Spaan and Marek Petrik for providing algorithmic results and code respectively. Research supported in part by AFOSR MURI project #FA9550-091-0538.